\documentclass[runningheads]{llncs}
\usepackage[T1]{fontenc}
\usepackage{graphicx,verbatim}

\usepackage{multirow}
\usepackage[dvipsnames]{xcolor}
\usepackage{amsmath,amssymb,amsfonts}
\usepackage{makecell}
\begin{document}
\title{EchoTracker2: Enhancing Myocardial Point Tracking by Modeling Local Motion}
\titlerunning{EchoTracker2: Myocardial Point Tracking by Modeling Local Motion}
%

\author{Md Abulkalam Azad\inst{1,2}\orcidID{0009-0006-7177-4961} \and
Vegard Holmstrøm\inst{1} \and
John Nyberg\inst{1} \and
Lasse Lovstakken\inst{1} \and
Håvard Dalen\inst{1,2} \and
Bjørnar Grenne\inst{1,2} \and
Andreas {\O}stvik\inst{1,2,3}\orcidID{0000-0003-3895-2683}}

\authorrunning{M. A. Azad et al.}

\institute{Norwegian University of Science and Technology, Trondheim, Norway \and
Clinic of Cardiology, St. Olavs Hospital, Trondheim, Norway \and
SINTEF Digital, Trondheim, Norway\\
\email{\{md.a.azad,andreas.ostvik\}@ntnu.no}}

\maketitle              

\begin{abstract}
Myocardial point tracking (MPT) has recently emerged as a promising direction for motion estimation in echocardiography, driven by advances in general-purpose point tracking methods. However, myocardial motion fundamentally differs from motion encountered in natural videos, as it arises from physiologically constrained deformation that is spatially and temporally continuous throughout the cardiac cycle. Consequently, motion trajectories typically remain locally confined despite substantial tissue deformation. Motivated by these properties, we revisit the architectural design for MPT and find that coarse initialization in commonly used two-stage coarse-to-fine architectures may be unnecessary in this domain. In this work, we propose a fine-stage-only architecture, \textbf{EchoTracker2}, which enriches pixel-precise features with local spatiotemporal context and integrates them with long-range joint temporal reasoning for robust tracking. Experimental results across in-distribution, out-of-distribution (OOD), and public synthetic datasets show that our model improves position accuracy by $6.5\%$ and reduces median trajectory error by $12.2\%$ relative to a domain-specific state-of-the-art (SOTA) model. Compared to the best general-purpose point tracking method, the improvements are $2.0\%$ and $5.3\%$, respectively. Moreover, EchoTracker2 shows better agreement with expert-derived global longitudinal strain (GLS) and enhances test-rest reproducibility. Source code will be available at: https://github.com/riponazad/ptecho.

\keywords{Point tracking  \and Motion estimation \and Echocardiography \and Ultrasound \and Strain imaging.}

\end{abstract}
\section{Introduction}
Tracking in general computer vision tasks must account for substantial variability in motion patterns arising from object movement, camera motion, occlusions and scene changes. Motion can be rigid or non-rigid, temporally inconsistent, making long-range correspondence inherently ambiguous. In contrast, myocardial motion in echocardiography is governed by physiological deformation that is spatially and temporally coherent, continuous, and constrained by anatomical structure~\cite{buckberg2008structure}. Although ultrasound imaging presents challenges such as speckle noise, low contrast, and view-dependent artifacts, tissue motion itself follows smooth and cyclic trajectories driven by cardiac contraction and relaxation~\cite{d2007principles}. 

\begin{figure}[t]
    \centering
    \includegraphics[width=\linewidth]{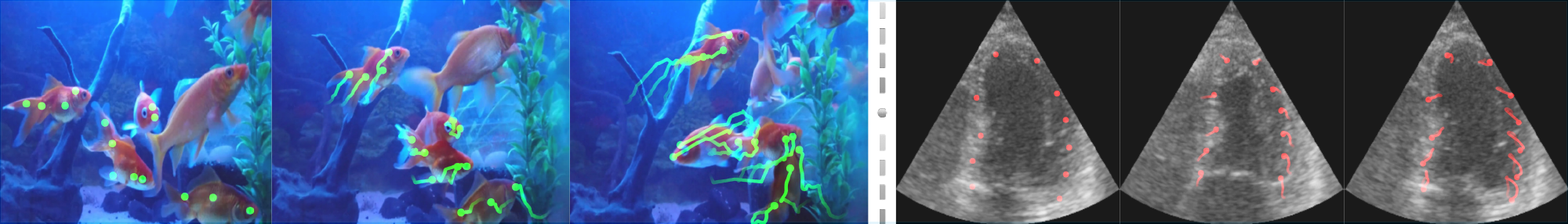}
    \\[-0.3cm]
    \caption{Motion trajectories of selected query points are shown on the first, middle, and last frames for a natural video (left, TAP-Vid DAVIS~\cite{doersch2022tap}) and a cardiac video (right).}
    \label{fig:local_motion}
    ~\\[-0.5cm]
\end{figure}

As in Fig.~\ref{fig:local_motion}, queried points in natural videos may move abruptly in arbitrary directions with large displacements, whereas motion in echocardiographic sequences is comparatively structured and remains confined to a local spatial neighborhood across frames. Importantly, this behavior arises from biomechanical constraints of myocardial tissue rather than acquisition conditions alone~\cite{claus2015tissue}. Continuity of deformation across neighboring regions and periodic contraction-relaxation dynamics enforce strong spatial and temporal priors on motion trajectories throughout the cardiac cycle~\cite{zhu2010coupled}. Consequently, while general-purpose tracking methods must accommodate unconstrained global motion, myocardial motion estimation can instead exploit these structural constraints.  Motivated by this, we hypothesize that focusing solely on local motion is sufficient for robust tracking in echocardiography, while enabling improved pixel-level accuracy that is critical for downstream functional measurements such as strain.

In this work, we propose \textbf{EchoTracker2}, a single-stage architecture for fine-grained myocardial point tracking across a full cardiac cycle that explicitly leverages locally constrained motion. In contrast to existing coarse-to-fine approaches, such as the ones proposed by Azad et al.~\cite{azad2024echotracker} and Chernyshov et al.~\cite{chernyshov2025myo}, our method models correspondence within a confined spatial neighborhood and integrates local spatiotemporal feature representations with joint temporal reasoning across neighboring trajectories. Experimental results demonstrate improved tracking performance and generalizability across in-distribution, OOD, and synthetic datasets. Furthermore, EchoTracker2 improves agreement and reproducibility in downstream GLS measurements compared to existing methods.
\section{Methods}
Given a cardiac video $\mathcal{V} \in \mathbb{R}^{T\times H \times W \times C}$ and a set of $N$ query points $\mathcal{P} = \{(x_i^{t_q}, y_i^{t_q}) \in \mathbb{R}^2 \mid i = 0, \dots, N-1\}$ selected on the myocardium from an arbitrary frame $t_q \in \{0, \dots, T-1\}$, the objective is to estimate the point trajectories $\mathcal{T} = \{(x_i^t, y_i^t) \mid i = 0, \dots, N-1\,;\, t = 0 \dots, T-1\}$ throughout the cardiac cycle.

\subsection{EchoTracker2}
\begin{figure}[h]
    \centering
    \includegraphics[width=\linewidth]{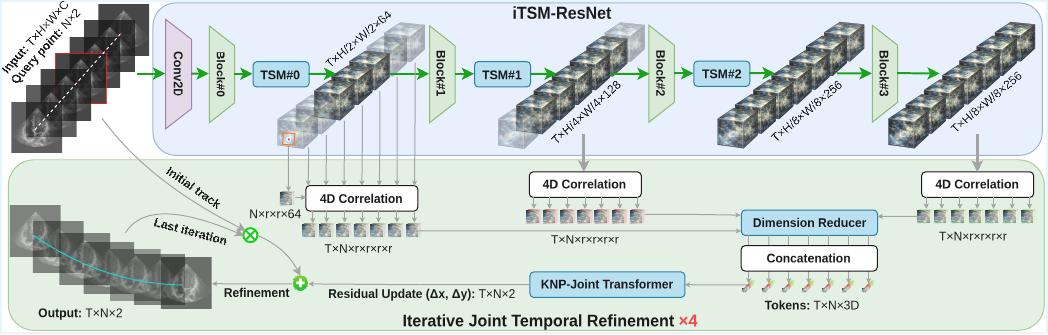}
    \\[-0.3cm]
    \caption{The architecture of EchoTracker2 and the computational flow for tracking a single point across ultrasound frames. Within each TSM block, temporal context is aggregated from adjacent frames, progressively expanding the temporal receptive field toward deeper blocks (indicated by the transparency of feature volumes). Although shown only for the middle frame (red border), the same process applies to all frames. The final trajectories are obtained through four iterations of joint temporal refinements.}
    \label{fig:echotracker2}
    ~\\[-0.7cm]
\end{figure}

We design EchoTracker2 to model the locally constrained deformation that characterizes myocardial motion. Unlike general-purpose tracking methods, which rely on coarse initialization for global motion to accommodate large and unpredictable displacements, myocardial trajectories are governed by continuous tissue deformation and exhibit limited inter-frame displacement. This allows correspondence estimation to be performed within a confined spatial neighborhood without requiring global correlation. Hence, we eliminate the commonly used coarse initialization, focus on learning pixel-precise local spatiotemporal representations, and refine trajectories through joint temporal reasoning across neighboring points.
An overview of EchoTracker2 is shown in Fig.~\ref{fig:echotracker2}, and its components are described below.

\subsubsection{Feature extraction.} 
Following Doersch et al.~\cite{doersch2022tap,doersch2023tapir}, we adopt the temporal shift module (TSM)-ResNet~\cite{lin2019tsm} to extract temporally-aware features. In prior work, TSMs are applied only to the deepest ResNet block, despite multi-scale correlation being computed across features from all blocks. This introduces a mismatch between spatial resolution and temporal context, as shallower feature maps used for correspondence estimation remain temporally independent. In the context of myocardial motion, where correspondence is estimated within a confined spatial neighborhood and trajectories evolve smoothly over time, feature representations at all scales should encode local temporal information. We therefore integrate lightweight TSM layers between each ResNet block, yielding a temporally-enhanced backbone referred to as \textbf{iTSM-ResNet}.
Each TSM aggregates context from adjacent frames, progressively expanding the temporal receptive field to 7 frames (3 backward and 3 forward) at the deepest block, as illustrated in Fig.~\ref{fig:echotracker2}. Let $\mathcal{F}^{(0)} \in \mathbb{R}^{T \times \frac{H}{k^{(0)}} \times \frac{W}{k^{(0)}} \times d^{(0)}}$ denote the intermediate frame features before the first $(\ell=0)$ ResNet block $\mathcal{B}_0$, where $k^{(0)}$ is the spatial downsampling factor and $d^{(0)}$ is the feature dimension. The feature extraction process of iTSM-ResNet can then be summarized as;
\begin{equation}
\mathcal{F}^{(\ell+1)} =
\begin{cases}
\mathcal{TSM}_\ell\big(\mathcal{B}_\ell(\mathcal{F}^{(\ell)})\big), & \ell \in \{0,1,2\} \\
\mathcal{B}_\ell(\mathcal{F}^{(\ell)}), & \ell = 3
\end{cases}
\end{equation}

\subsubsection{Correlation computation.} 
We estimate correspondence by computing feature correlation within a local spatial neighborhood. While 2D correlation is widely used, we adopt 4D correlation~\cite{cho2024local} due to its robustness to matching ambiguity, a key advantage for ultrasound imaging. As illustrated for a single point at one feature level in Fig.~\ref{fig:echotracker2}, query features $\mathcal{Q}^{\ell} \in \mathbb{R}^{N \times r \times r \times d^{(\ell)}}$ are bilinearly sampled for $N$ points from the query frame within an $(r \times r)$ window at level $\ell$. Corresponding frame features $\mathcal{F}_c^{(\ell)} \in \mathbb{R}^{T \times N \times r \times r \times d^{(\ell)}}$ are sampled at the same spatial resolution $(r \times r)$ along the previously estimated trajectories (in the first iteration, point locations from the query frame are propagated to all frames). All-pair cosine similarity is computed between $\mathcal{Q}^{(\ell)}$ and $\mathcal{F}_c^{(\ell)}$ to obtain high-dimensional correlation maps $\mathcal{C}^{(\ell)} \in \mathbb{R}^{T \times N \times r \times r \times r \times r}$. These maps are subsequently encoded into a compact representation of dimensionality $D$ using a learnable encoder~\cite{cho2024local} and concatenated across three feature scales $(\mathcal{F}^{(1)}, \mathcal{F}^{(2)}, \mathcal{F}^{(4)})$, yielding multi-scale correlation tokens $\mathcal{C} \in \mathbb{R}^{T \times N \times 3D}$.

\subsubsection{Joint Temporal refinement.}
Although features enriched with local spatiotemporal context are often sufficient for short-term tracking, the accuracy may degrade in the presence of out-of-plane motion or image artifacts. Besides, myocardial deformation exhibits spatially coherent motion among neighboring regions, and the coherency differs progressively for distant regions. Thus, nearby points have stronger motion dependency than distant ones. To account for this property, we extend the transformer of Cho et al.~\cite{cho2024local} by applying multi-head self-attention~\cite{vaswani2017attention} both along trajectories and across $K$ nearest points simultaneously for joint temporal reasoning, referred to as the \textbf{KNP-Joint} transformer (Fig.~\ref{fig:echotracker2}). The module takes the correlation tokens $\mathcal{C}$ as input and predicts residual trajectory updates $(\Delta x, \Delta y) \in \mathbb{R}^{T \times N \times 2}$. The final refined trajectories are obtained iteratively as $\mathcal{T}^{(m)} = \mathcal{T}^{(m-1)} + (\Delta x, \Delta y)^{(m)}$ after $m=4$ iterations.

\section{Experiments}\label{expr}
\subsection{Datasets}
We use real-world in-house echocardiography datasets collected with GE Vivid E95 and E9 systems with ethical approval and patient consent (Table~\ref{tab_ds}). Each video captures a full cardiac cycle  from end-diastole to the subsequent end-diastole frame. For each sequence, myocardial trajectories are obtained by tracking $N$ sparse points along the left or right ventricular wall~(LV or RV). These trajectories are generated using a semi-supervised tracking algorithm and subsequently refined through quality-controlled manual verification by experienced clinical experts (>20 years). On average, each video contains 50-80 tracked myocardial points over 50-150 frames per cardiac cycle.

All models, including SOTA baselines and EchoTracker2, are trained or fine-tuned exclusively on the Training dataset and evaluated on separate in-house and public datasets. The In-distribution dataset contains LV motion from both healthy subjects and patients with cardiac pathology acquired from standard views. The OOD RV dataset consists of RV-focused acquisitions from healthy subjects and represents a distribution shift in both anatomy and imaging view. We additionally evaluate generalizability on the publicly available synthetic CAMUS dataset~\cite{evain2022motion}, which provides myocardial trajectories sampled across multiple endocardial-to-epicardial contours, in contrast to the single mid-wall trajectory used during training. Finally, the Agreement and Reproducibility datasets are used to assess GLS agreement with experts and test–retest stability, respectively. The latter consists of repeated acquisitions from the same patient obtained within the same session under comparable physiological conditions.

\begin{table}[h]
\footnotesize
\centering
\caption{Overview of the echocardiographic datasets.}
\label{tab_ds}
\setlength{\tabcolsep}{6pt} 
\begin{tabular}{lcccc}
\\[-0.5cm]
\hline
\textbf{Dataset} & \textbf{Patients} & \textbf{Videos} & \textbf{Pathology} & \textbf{Views}\\
\hline
 Training & 643 & 1922 & Healthy & A4C\,\textperiodcentered\,A2C\,\textperiodcentered\,A3C \\
 In-distribution & 371 & 1112 & Mixed & A4C\,\textperiodcentered\,A2C\,\textperiodcentered\,A3C \\
OOD RV & 1254 & 1331 & Healthy & A4C (RV-focused) \\
CAMUS & 98 & 196 & Undefined & A4C\\[0.1cm]
Agreement & 203 & 609 & Mixed & A4C\,\textperiodcentered\,A2C\,\textperiodcentered\,A3C  \\
Reproducibility & 127 & 381 & Mixed & A4C\,\textperiodcentered\,A2C\,\textperiodcentered\,A3C  \\
\hline
\end{tabular}
~\\[0.05cm]
\raggedright
{\scriptsize Mixed: Healthy subjects and patients with heart failure, myocardial infarction or atrial fibrillation. A4C: Apical four-chamber, A2C: Apical two-chamber, A3C: Apical three-chamber/long-axis}
\end{table}

\subsection{Evaluation Metrics}
Tracking performance is commonly evaluated as the average percentage of tracked points  that remain within a spatial tolerance $x$ of their ground-truth locations along the trajectories, averaged over multiple pixel thresholds ($<$$\delta^x_{avg}$)~\cite{doersch2022tap,doersch2023tapir}. Standard protocols include tolerances up to $x=16$ pixels, however, such thresholds are less relevant for clinical strain estimation where small tracking errors may substantially affect deformation measurements. We therefore report $<$$\delta^x_{avg}$ only for $x\in \{1, 2, 4\}$ pixels to better reflect clinically meaningful tracking accuracy. In addition, we report the median trajectory error (MTE) in pixels and the average inference time (AIT) in seconds per video. Distances are computed using the $\mathcal{L}_1$ distance at a spatial resolution of $256\times256$ pixels.

Clinical evaluation is performed using peak GLS, which quantifies the relative change of the longitudinal ventricular length from end-diastole to
minimum ventricular length. Agreement with expert-derived measurements is assessed using the mean and standard deviation of differences, while reproducibility is evaluated using the mean absolute difference (MAD) and coefficient of variation (CV).

\subsection{Implementation details}
All SOTA models (based on their open-source implementations) and EchoTracker2 are implemented in PyTorch. For training and fine-tuning, we adopt the impartial motion augmentation strategy proposed in~\cite{azad2025taming}. Each model is trained for 100 epochs (8,620 augmented clips, each containing 36 frames with 40 query points; batch size 4). Training each model requires approximately one week on a single A6000 (48GB) GPU. The training hyperparameters (learning rate = $5 \times 10^{-4}$, one-cycle scheduler, and AdamW optimizer) are kept identical across models to ensure fair comparison. All evaluations in Section~\ref{res} are performed on a single RTX 3090 (24GB) GPU. For supervision, we compute the $\mathcal{L}_1$ distance between the reference trajectories $\mathcal{T}^{gt}$ and the estimated trajectories $\mathcal{T}$, and apply exponential weighting ($\gamma = 0.8$) over the $m$ refinement iterations as
\begin{equation}
    \mathcal{L}oss = \frac{1}{m}\sum_{i=1}^{m}\gamma^{m-i}\mathcal{L}_1(\mathcal{T}^{(i)}, \mathcal{T}^{gt})
\end{equation}

\subsection{Ablation Study}
We conduct ablation studies to evaluate the impact of key architectural design choices. All variants are trained for 50 epochs on 200 sequences (36 frames each) using identical hyperparameters and validated on a held-out subset of 40 patients from the In-distribution dataset to ensure consistent comparison.

\subsubsection{Local Spatial Correlation.}
Since EchoTracker2 restricts correspondence estimation to a local spatial neighborhood, the correlation window must be sufficiently large to capture typical inter-frame displacements. Although several SOTA methods~\cite{azad2024echotracker,cho2024local,karaev2025cotracker3} have used a $7\times7$ window, the optimal size depends on input resolution, downsampling factor, and motion characteristics of the imaging domain. We therefore evaluate four window sizes for 4D correlation, as shown in Table~\ref{tab:ablation_merged}.a. A window size of $9\times9$ provides the best trade-off between tracking accuracy and computational cost for EchoTracker2.

\subsubsection{Local Temporal Feature Enrichment.} 
We explore several strategies to enrich frame features with neighboring temporal context (Table~\ref{tab:ablation_merged}.b). 
An efficient approach is to fuse precomputed local spatial features from adjacent frames prior to correlation, either through addition or concatenation, similar to~\cite{azad2024echotracker,zheng2023pointodyssey}. Alternatively, temporal reasoning can be incorporated directly into the backbone using learnable modules such as TSM-ResNet~\cite{doersch2023tapir}. We implement TSM in two ways: blockwise (bTSM-ResNet), where features from each block pass through a single-layer TSM without altering the original data flow, and injection (iTSM-ResNet), where a TSM is inserted after each block so its output is propagated to the subsequent block.

\subsubsection{Long-range Temporal Reasoning.}
We first extend the transformer of Cho et al.~\cite{cho2024local} with self-attention across all points for full joint reasoning and compare it to the cross-attention-based transformer of Karaev et al.~\cite{karaev2025cotracker3}. Our final variant, KNP-Joint temporal transformer, performs joint reasoning over the $k=10$ nearest neighboring trajectories and achieves the best performance (Table~\ref{tab:ablation_merged}.c).

\begin{table}[h]
    ~\\[-0.5cm]
    \footnotesize
    \centering
    \caption{Ablation studies. Impact of (a) different spatial window sizes for 4D correlation, (b) local temporal feature enrichment, (c) joint temporal reasoning.}
    \setlength{\tabcolsep}{1.pt}
    \begin{tabular}{|c|ccccc|ccccc|}
    \\[-0.5cm]
    \hline
    \multirow{2}{*}{\textbf{Configuration}} & \multicolumn{5}{c|}{\textbf{In-distribution}} & \multicolumn{5}{c|}{\textbf{OOD RV}} \\
    \cline{2-11}
    & $<$$\delta^1$$\uparrow$ & $<$$\delta^2$$\uparrow$ & $<$$\delta^4$$\uparrow$ & $MTE$$\downarrow$ & $AIT$$\downarrow$ & $<$$\delta^1$$\uparrow$ & $<$$\delta^2$$\uparrow$ & $<$$\delta^4$$\uparrow$ & $MTE$$\downarrow$ & $AIT$$\downarrow$\\
    \hline
    \multicolumn{11}{|c|}{\textit{(a) Spatial window size}} \\
    \hline
    $5\times5$ & 29.99 & 54.56 & 80.78 & 2.24 & \textbf{0.17} & 20.65 & 38.97 & 62.59 & 3.73 & \textbf{0.19}\\
    $7\times7$ & 30.53 & 54.65 & 80.93 & 2.22 & 0.21 & 21.42 & 39.92 & 63.38 & 3.70 & 0.21\\
    $9\times9$ & \textbf{31.82} & \textbf{55.83} & \textbf{81.42} & \textbf{2.17} & 0.27 & \textbf{22.34} & \textbf{41.48} & \textbf{65.01} & \textbf{3.54} & 0.28\\
    $11$$\times$$11$ & 30.88 & 55.04 & 81.15 & 2.21 & 0.35 & 21.04 & 39.94 & 63.98 & 3.61 & 0.36\\
    \hline
    \multicolumn{11}{|c|}{\textit{(b) Temporal feature enrichment}} \\
    \hline
    BL$+$Fusion(add) & 32.44 & 56.40 & 81.57 & 2.14 & \textbf{0.27} & 22.40 & 40.75 & 63.68 & 3.59 & \textbf{0.29}\\
    BL$+$Fusion (cat) & 32.82 & 57.01 & \textbf{82.52} & 2.10 & \textbf{0.27} & 22.83 & 41.91 & 65.65 & 3.44 & 0.30 \\
    bTSM-ResNet & 32.98 & 57.14 & 82.50 & 2.10 & 0.33 & \textbf{23.29} & \textbf{42.75} & \textbf{66.24} & 3.45 & 0.37 \\
    iTSM-ResNet & \textbf{33.31} & \textbf{57.40} & 82.46 & \textbf{2.08} & 0.32 & 23.20 & 42.27 & 65.85 & \textbf{3.43} & 0.36 \\
    \hline
    \multicolumn{11}{|c|}{\textit{(c) Joint reasoning strategy}} \\
    \hline
    Joint-Reasoning & 33.12 & 57.29 & 82.39 & 2.09 & 0.34 & 23.16 & 42.19 & 65.51 & 3.50 & 0.37\\
    Cross-Attention & 32.70 & 56.56 & 81.76 & 2.13 & 0.29 & 23.05 & 42.13 & 65.32 & 3.49 & 0.32 \\
    KNP-Joint & \textbf{33.31} & \textbf{57.40} & \textbf{82.46} & \textbf{2.08} & \textbf{0.32} & \textbf{23.20} & \textbf{42.27} & \textbf{65.85} & \textbf{3.43} & 0.36 \\
    \hline
    \end{tabular}
    \label{tab:ablation_merged}
\end{table}
\section{Results}\label{res}
\subsubsection{Local Motion in Echocardiography.}
Table~\ref{tab:wo_init} shows that removing the initialization stage from EchoTracker and LocoTrack results in negligible differences in tracking performance, despite both models being trained with it. This suggests that modeling local motion alone is sufficient and reduces computational cost.

\begin{table}[h]
    \footnotesize
    \centering
    \caption{Performance comparison of tracking and computational cost without initialization stage on In-distribution dataset.}
    \setlength{\tabcolsep}{3.5pt} 
    \begin{tabular}{c|ccccc|ccccc}
    \\[-0.5cm]
    \hline
    \multirow{2}{*}{\makecell{\textbf{Initial} \\ \textbf{Stage?}}} & \multicolumn{5}{c|}{\textbf{EchoTracker~\cite{azad2024echotracker}}} & \multicolumn{5}{c}{\textbf{LocoTrack~\cite{cho2024local}}} \\
    \cline{2-11}
    & $<$$\delta^1$$\uparrow$ & $<$$\delta^2$$\uparrow$ & $<$$\delta^4$$\uparrow$ & $MTE$$\downarrow$ & $AIT$$\downarrow$ & $<$$\delta^1$$\uparrow$ & $<$$\delta^2$$\uparrow$ & $<$$\delta^4$$\uparrow$ & $MTE$$\downarrow$ & $AIT$$\downarrow$\\
    \hline
    Yes & 37.07 & 64.01 & 88.73 & 1.82 & 0.15 & 39.95 & 65.70 & 89.19 & 1.73 & 0.23\\
    No & 37.54 & 63.89 & 88.30 & 1.83 & 0.12 & 39.55 & 65.40 & 89.23 & 1.73 & N\!/\!A\\
    \hline
    \end{tabular}
    \label{tab:wo_init}
    \\[-0.3cm]
\end{table}

\subsubsection{Tracking performance.}
To benchmark EchoTracker2, we compare its performance with leading SOTA models from both the general-purpose and echocardiography domains (Table~\ref{tab:res}). Although pretrained PIPs++, LocoTrack (base), and CoTracker3 (offline) were trained on large-scale general video datasets~\cite{cho2024local,karaev2025cotracker3,zheng2023pointodyssey} and further finetuned on our training dataset, EchoTracker2, trained only echocardiography videos for 100 epochs, outperforms them across In-distribution, OOD RV, and synthetic CAMUS datasets. Averaged across datasets, EchoTracker2 improves position accuracy by $6.52\%$ and reduces MTE by $12.22\%$ relative to EchoTracker, and surpasses the best performing general-purpose point tracking model, LocoTrack, by $2.02\%$ and $5.28\%$, respectively. Some qualitative tracking results are attached in the supplement.

\begin{table}[h]
\footnotesize
    \centering
    \caption{Overall tracking performance of the models on In-distribution, OOD RV, and synthetic CAMUS datasets.}
    \setlength{\tabcolsep}{2.2pt} 
    \begin{tabular}{l|ccc|ccc|ccc}
    \\[-0.5cm]
    \hline 
    \multirow{2}{40pt}{\textbf{Methods}} & \multicolumn{3}{c|}{\textbf{In-distribution}} & \multicolumn{3}{c|}{\textbf{OOD RV}} & \multicolumn{3}{c}{\textbf{CAMUS}~\cite{evain2022motion}}\\
    \cline{2-10}
    & $<$$\delta^x_{avg}$$\uparrow$ & MTE$\downarrow$ & AIT$\downarrow$ & $<$$\delta^x_{avg}$$\uparrow$ & MTE$\downarrow$ & AIT$\downarrow$ & $<$$\delta^x$$\uparrow$ & MTE$\downarrow$ & AIT$\downarrow$\\
    \hline
    PIPs++~\cite{zheng2023pointodyssey} & 57.40 & 2.06 & \textbf{0.14} & 41.78 & 3.40 & \textbf{0.16} & 54.90 & 2.25 & \textbf{0.17}\\
    SpeckNet~\cite{azad2025taming} & 59.56 & 1.97 & 0.20 & 43.66 & 3.53 & 0.26 & 64.24 & 1.81 & 0.26\\
    EchoTracker~\cite{azad2024echotracker} & 63.27 & 1.82 & 0.15 & 46.63 & 3.17 & 0.19 & 70.33 & 1.58 & 0.18\\
    CoTracker3~\cite{karaev2025cotracker3} & 64.62 & 1.74 & 0.55 & 47.60 & 3.02 & 0.64 & 71.96 & 1.43 & 0.70\\
    LocoTrack~\cite{cho2024local} & 64.95 & 1.73 & 0.23 & 49.34 & 2.93 & 0.24 & 73.71 & 1.43 & 0.29\\
    EchoTracker2(ours) & \textbf{66.94} & \textbf{1.63} & 0.33 & \textbf{49.84} & \textbf{2.84} & 0.37 & \textbf{75.17} & \textbf{1.33} & 0.47 \\
    \hline
    \end{tabular}
    \label{tab:res}
\end{table}

\subsubsection{Evaluation of GLS agreement and test–retest reproducibility.}
Table~\ref{tab_clinic_perf} summarizes the GLS obtained using EchoTracker2 in comparison with expert observers, the best-performing general-purpose tracking model, and the domain-specific EchoTracker baseline. Agreement metrics reflect AI-to-observer variability, while the expert observer row reports inter-observer variability. Reproducibility is evaluated in a test–retest setting, with the expert observer row indicating intra-observer variability. EchoTracker2 demonstrates improved agreement with expert measurements compared to existing tracking methods, while also reducing the coefficient of variation in test–retest reproducibility.

\begin{table}[h]
\footnotesize
    \centering
    \caption{GLS agreement with expert observers and test-retest reproducibility.}
    \setlength{\tabcolsep}{3.8pt} 
    \begin{tabular}{lccccccccc}
    \\[-0.5cm]
    \hline
      & \multicolumn{4}{c}{\textbf{Agreement}} & & \multicolumn{3}{c}{\textbf{Reproducibility}} \\
      \cline{2-5} \cline{7-9}
      \textbf{Method} & Mean$\pm$SD & $\mu$$\downarrow$ & $\sigma$$\downarrow$ & MAD$\downarrow$ & & Mean$\pm$SD & MAD$\downarrow$ & CV$\downarrow$ \\
      \hline
      Expert observers & $-$17.7$\pm$3.4 & 0.58 & 1.80 & 1.44 & & $-$17.3$\pm$3.5 & 1.29 & 6.8 \\[0.1cm]
      EchoTracker & $-$15.9$\pm$3.2 & 1.72 & 1.47 & 1.95 & & $-$15.5$\pm$3.3 & \textbf{1.21} & 6.9 \\
      LocoTrack & $-$16.0$\pm$3.2 & 1.60 & 1.46 & 1.86 & & $-$15.7$\pm$3.3 & 1.30 & 7.2 \\
      EchoTracker2 (ours) & $-$16.9$\pm$3.3 & \textbf{0.76} & \textbf{1.44} & \textbf{1.41} & & $-$16.5$\pm$3.4 & 1.24 & \textbf{6.6} \\
      \hline
      \end{tabular}
      \label{tab_clinic_perf}
      \raggedright
      {\scriptsize $\mu$: mean difference (\%), $\sigma$: standard deviation of differences (\%), MAD: mean absolute difference (\%), CV: coefficient of variation (\%). Agreement columns show AI-vs-observer metrics; expert observer row shows inter-observer variability. Reproducibility expert row shows intraobserver.}
      \\[-0.4cm]
\end{table}
\section{Conclusion}
We present \textbf{EchoTracker2}, a fine-stage-only MPT model, that enriches features with local spatiotemporal context and leverages 4D correlation for precise tracking. Our joint-temporal refinement across neighboring points improves robustness to out-of-plane motion and ultrasound artifacts. EchoTracker2 outperforms existing SOTA methods and demonstrates improved generalizability across in-distribution, OOD and public synthetic data. Furthermore, it improves agreement with expert-derived GLS measurements, while maintaining strong test-retest reproducibility. These results suggest that domain-specific motion modeling may enhance the reliability of deformation-based myocardial function assessment in routine echocardiographic practice.

%
%
%
\bibliographystyle{splncs04}
\bibliography{ref}
%




\end{document}